\documentclass{article}

\usepackage{arxiv}

\usepackage[utf8]{inputenc} 
\usepackage[T1]{fontenc}    
\usepackage{hyperref}       
\usepackage{url}            
\usepackage{booktabs}       
\usepackage{amsfonts}       
\usepackage{nicefrac}       
\usepackage{microtype}      
\usepackage{url}
\usepackage{svg}
\usepackage{subfigure}
\usepackage{graphicx}
\graphicspath{ {./images/} }
\usepackage{caption}
\usepackage[section]{placeins}

\usepackage{amsmath}
\DeclareMathOperator*{\argmax}{argmax}

\usepackage[ruled,vlined,linesnumbered,noresetcount]{algorithm2e}
\makeatletter
\newcommand{\AlgoResetCount}{\renewcommand{\@ResetCounterIfNeeded}{\setcounter{AlgoLine}{0}}}
\newcommand{\AlgoNoResetCount}{\renewcommand{\@ResetCounterIfNeeded}{}}
\newcounter{AlgoSavedLineCount}

\makeatother

\title{Random Search as a Baseline for Sparse Neural Network Architecture Search}

\date{} 					

\author{Rezsa Farahani\\
	Google, USA\\
}

\begin{document}
\maketitle

\begin{abstract}

\end{abstract}
Sparse neural networks have shown similar or better generalization performance than their dense counterparts while having higher parameter efficiency. This has motivated a number of works to learn or search for high performing sparse networks. While reports of task performance or efficiency gains are impressive, standard baselines are lacking leading to poor comparability and unreliable reproducibility across methods. In this work, we propose Random Search as a baseline algorithm for finding good sparse configurations and study its performance. We apply Random Search on the node space of an overparameterized network with the goal of finding better initialized sparse sub-networks that are positioned more advantageously in the loss landscape. We record the post-training performances of the found sparse networks and at various levels of sparsity, and compare against both their fully connected parent networks and random sparse configurations at the same sparsity levels. First, we demonstrate performance at different levels of sparsity and highlight that a significant level of performance can still be preserved even when the network is highly sparse. Second, we observe that for this sparse architecture search task, initialized sparse networks found by Random Search neither perform better nor converge more efficiently than their random counterparts. Thus we conclude that Random Search may be viewed as a reasonable neutral baseline for sparsity search methods.


\section{Introduction}
Overparameterized neural networks are loosely characterized as networks that have a very high fitting (or memorization) capacity with respect to their training data. Although capable of memorization of their training data, these networks intriguingly achieve very low test error close to their training error rates \cite{10.1145/3446776, neyshabur2018the}. Meanwhile, sparse neural networks have shown similar or better generalization performance than their dense counterparts while having higher parameter efficiency \cite{10.5555/3546258.3546499}. With increasing availability of hardware and software that support sparse computational operations \cite{8014795, 9065428}, there has been a growing interest in finding sparse sub-networks within large overparameterized models to either improve generalization performance or to gain computational efficiency at the same performance level \cite{DBLP:journals/corr/abs-1907-04840, 9156601, DBLP:journals/corr/Graham14, 10.5555/3546258.3546499}.

Earlier works on creating efficient sparse sub-networks include the now popular pruning technique \cite{10.5555/2969830.2969903}. These were motivated by the desire to achieve compute efficiency in resource constraint applications by finding smaller networks within a larger network space without losing task performance quality \cite{blalock2020state}. The original pruning technique involves fully training a larger network on some task, discarding the task-irrelevant connections, and then fine-tuning the remaining sparse sub-network on the task to achieve the a level of performance near that of the larger network. Connections were originally pruned based on loss Hessians \cite{10.5555/2969830.2969903, hassibi1993optimal}. Later on, other techniques were proposed such as the removal of weak connections \cite{han2015learning} based on weight value thresholds.

Frankle \& Carbin \cite{frankle1810lottery} proposed a method for more efficient training of overparameterized networks. In this method, named \textit{Iterative Magnitude Pruning with Rewinding (IMP)}, sub-networks of larger networks were found that could be trained to reach a similar level of performance to that of their larger superset network. The IMP process requires training a full network to convergence, pruning the network by removing or masking connections with weaker weights, and rewinding the weight values of the stronger connections to their initial values before re-training the sparse sub-network. The original IMP method requires resetting the sub-network weights to their original values, however, other studies have argued that this resetting may not be needed \cite{liu2018rethinking, zhou2019deconstructing} which relaxes the need to preserve the non-pruned weights. Additionally, Liu et al. \cite{liu2018rethinking} show that both full training of the overparameterized model for pruning and fine-tuning sub-networks are also not required for certain architectures.

Other work in this area include studies on weight-agnostic neural networks by Gaier \& Ha \cite{gaier2019weight}, which by only evolving the network architecture and not training weights were able to achieve high accuracy values on the MNIST image classification task. In another line of work by Zhou et al. \cite{zhou2019deconstructing} and similarly in a follow up by Ramanujan et al. \cite{ramanujan2019s}, a \textit{super mask} is trained (with SGD) to score connection weights. A set of connections are then selected according to their weights' scores for finding effective sparse connections in an overparameterized network. Most interestingly, Ramanujan et al. \cite{ramanujan2019s} demonstrated that the found sub-networks in randomly initialized overparameterized networks achieve good performance on certain tasks without making any updates to the original weight values. Finally, reinforcement learning (RL) methods have been applied for pruning networks with various degrees of success \cite{DBLP:journals/corr/abs-2007-04756}.

In this work, we aim to develop a baseline search method for finding desirable sparse configurations within overparameterized neural networks prior to full training. This baseline should be reasonably naive but also not a trivial one. We desire sparse networks that exhibit smaller deterioration in task performance compared to a full network, versus purely random sparse configurations at the same level of sparsity. If any search method should arrive at better initial configurations prior to gradient-based training as compared to our baseline, then there is support that the successful method is at least better than a baseline that is not deemed trivial. We propose our baseline for a class of methods that perform connection architecture search on the edges or nodes of an untrained larger overparameterized network with random weight initialization. They aim to find good sparsely connected sub-networks prior to training which would then be normally trained to convergence e.g. with stochastic gradient descent. Our proposed baseline is intended to help measure progress on sparsity search methods, and to control for randomness in initialization and other random factors that may result in fleeting success and non-reproducible results\cite{frankle2021pruning}. Some works have proposed sets of benchmark tasks for analyzing competing sparse networks, hoping to bring a more rigorous evaluation methodology to the field \cite{liu2023sparsity}. Our work contributes a baseline algorithm for sparsity search to this front.

Most algorithms that search for effective sub-networks accomplish this by either pruning weak connections or by learning connection masks. In our approach towards establishing a baseline method, we use \textit{Random Search} for finding good initialized sparse sub-networks within the larger network. We name this approach \textit{Weedout}.

Weedout involves executing a simple Random Search process as the most basic directed search method among the more elaborate \textit{Random Search Heuristics} \cite{Auger_Doerr_2011} methods. In effect, it is a \textit{Neural Architecture Search} \cite{10.5555/3304415.3304617, JMLR:v20:18-598} process directed on the connection space of an overparameterized network. Random Search Heuristics (RSH), also known as \textit{Population-Based} search methods or \textit{Evolutionary Computation}, has a long record of being successfully applied to black-box optimization and machine learning problems. Recently, such methods have also been applied to large-scale Neural Architecture Search with success \cite{10.5555/3304415.3304617, liu2018hierarchical, real2017large}. RSH methods scale well with computational resources, and are very amenable to parallel and distributed computation. Furthermore, Random Search is non-parametric, does not perform gradient-based learning, and lacks any inductive bias. On the downside, it is not sample efficient. But as a primitive evolutionary algorithm, it is often regarded as a baseline search method.

We combine this idea with findings from recent studies on the early stages of neural network training, which show that the early training phase involves the most dynamic and important computational activity through the SGD training process \cite{achille2018critical, gur2018gradient, golatkar2019time, Frankle2020The}. Additionally, several works have focused on finding good initializations prior to training \cite{mishkin2015all, Dauphin_Schoenholz_2019}. We thus focus the Weedout procedure on the early training phase of overparameterized networks with the intention of finding better initialized sub-networks by using early performance scores which saves us from spending significant full training resources.

\section{The Method}

The Weedout method entails finding desirable sparsely-connected networks within overparameterized networks among a number of possible connection configurations (population in connection-space), which exhibit better task performance potential among their cohort \textit{before} full gradient-based training and weight adjustments with SGD. To find these desirable sparse networks, we use Random Search to "Weedout" bad solutions and keep the better solutions that are more promising given their early performance on a given task.

The Weedout training process consists of two phases, (1) the Weedout phase, and (2) the sparse network learning phase. First, a fully connected overparameterized neural network is initialized with random weights. A number of sparsely-connected sub-networks are then sampled from this fully connected network forming a population of sparse networks. Once the population is established, Random Search is applied to the population for finding the best solution among this population. We call this phase the Weedout phase. After finding the best solution, the sparse network is then trained normally with SGD as part of a normal learning phase. The Weedout phase is further discussed below.

Generally, any RSH method may be applied for the purpose of finding a better sparse network among the candidate population. Suitability and effectiveness of any given RSH strategy depends on a variety of factors, such as the given learning task, the population count, number of iterations for the strategy, etc. Random Search is a basic strategy and suitable as a relatively naive baseline compared to other search heuristics such as \textit{Binary Tournament} which is commonly used in Evolutionary Computation applications \cite{DeJong_2016}. 

During the Weedout phase, Random Search (or any such RSH strategy) may be applied iteratively for selecting a number of winning solutions among a population and passing winners to the next iteration. Each iteration in RSH terminology is also known as a \textit{generation}. The selection process of the RSH strategy at each generation is based on a search signal called the \textit{fitness} value of a given solution, where \textit{fitter} sparse networks are selected and passed to the next generation under the heuristics of their particular RSH strategy. Concretely, we consider fitness to be the negative \textit{cost} on a batch of examples sampled from a validation holdout set that is intended for hyperparameter optimization. This is a similar approach to the strategy of model selection based on an output metric as used in pruning proposed by Li et al. \cite{li2017pruning}, instead of eliminating connections with weak weights as done in pruning methods. After several generations of selection of fitter solutions, or meeting some termination criteria such as convergence in the best fitness value, the Weedout phase is stopped and the best sparse network solution is taken for SGD training on the task training data. As in normal SGD training, generalization performance is measured on a test holdout set.

The process of creating random sparse sub-networks may be analogous to the creation of randomly connected networks such as in the work of Xie et al. \cite{xie2019exploring}. However, creation of random networks from an overparameterized network creates additional opportunities to benefit from tensor-favorable hardware and software platforms without requiring computationally expensive graph computing facilities.

\begin{algorithm}[H]
  \KwData{Data splits: \textit{$D_{train}$}, \textit{$D_{validation}$}, \textit{$D_{test}$}; Network population count: \textit{m}; Random Search Heuristic: $\mathnormal{RSH}$; Weight initialized fully connected network: $\mathfrak{N}$; A relative degree of sparsity: $\eta$.}
  \For{$i\gets1$ \KwTo m}{
    $\mathfrak{n}^i$ $\leftarrow$ \scshape{RandomSubNetwork}($\mathfrak{N}$, $\eta$) \;
  }
  \While{$\neg$ \scshape{TerminationCriteria}}{
    $batch_{validation}$ $\leftarrow$ $batch_{validation}$ $\sim$ \textit{$D_{validation}$} \;
    \For{$i\gets1$ \KwTo m}{
      $\mathfrak{n}_{fitness}^i$ $\leftarrow$ $\mathnormal{-}$ $Cost(\mathfrak{n}^i, batch_{validation})$ \;
    }
    $\mathfrak{n}$ $\leftarrow$ $\mathnormal{RSH_{selection}}(\mathfrak{n})$ \;
  }
  $\mathfrak{n}^*$ $\leftarrow$ $\argmax_\mathfrak{n}{(\mathfrak{n}_{fitness})}$ \;
  \scshape{TrainAndUpdateWeights}($\mathfrak{n}^*$ , \textit{$D_{train}$}) \;
  \scshape{EvaluateGeneralizationPerformance}($\mathfrak{n}^*$, \textit{$D_{test}$}) \;
  \caption{WEEDOUT}
\end{algorithm}

\section{Experiments}

We conduct experiments to study the application of Weedout on an image classification task. We measure both train and test classification accuracy of fully trained sub-networks found by Weedout at various levels of larger network sparsity. For control, we compare sparse networks found by Weedout versus a randomly drawn sparse network from the population. This network can be conceptually thought of as a network with "fixed dropout nodes". Several of the mentioned work \cite{han2015learning, frankle1810lottery, liu2018rethinking, zhou2019deconstructing,  gaier2019weight, ramanujan2019s, xie2019exploring} run experiments on the CIFAR-10 image classification task using a VGG-style \cite{simonyan2015a} convolutional network architecture. We also conduct our experiments on the CIFAR-10 task, but use the Wide ResNet architecture \cite{Zagoruyko2016WRN} which generally is more reproducible and exhibits higher performance on single GPU machines than the VGG architectures.

We induce sparsity and control it by a hyperparameter $\eta$, the \textit{sparsity ratio}, which specifies the ratio of inactive nodes or weights in the fully connected larger network. At each generational iteration, Random Search selects the fittest solution in the population (the configuration with the highest validation set accuracy) and passes it on to the next generation, while creating $(m - 1)$ new random solutions and adding them to the next generation population. This process passes the best found solution through to subsequent generations for competing against new random solutions. We set the Weedout termination criterion to be 5 generations of selection on the population. This hyperparameter can take on an arbitrary value with a higher generation horizon giving the Weedout process more opportunity for solution exploration. However, the longer the Weedout phase takes, the more compute resources are consumed for search which can diminish the resource savings that a potential sparse network can bring. We start the generational search with a population of 100 randomly sampled sub-networks, after which the fittest sparse network solution over all generations is selected. The final network is then trained with SGD and undergoes weight updates via back propagation. Population size is another hyperparameter by which higher values allow for a broader initial exploration of the objective (fitness) landscape at the cost of requiring more compute resources.

Each randomly created sparse network is initialized under a \textit{He Normal} weight initialization \cite{he2015delving} scheme. We attempt to increase the chances of performance success for the untrained networks based on earlier work arguing in favor of neural networks with random Gaussian weights as universal classifiers \cite{giryes2016deep}. The sub-networks are created with induced sparsity in both the last fully-connected layers and in the convolutional blocks of the network. We run training on an 8 core tensor processing unit platform with a global batch size of 1024 (per-core batch size of 128).

For these experiments, we induce sparsity by \textit{node sparsity}, i.e. by randomly deactivating a number of nodes from each layer of the network. Deactivation of nodes is by applying a Hadamard product between layer activation outputs and a random 0/1 mask distributed randomly based on~$\eta$. This has the same effect on the flow of input values and gradients through the network as in removing connections or nodes from the actual network compute graph. Depending on whether sparsity is induced by a mask or actual absence of connection/node, the compute performance implications may be different based on the compute substrate and particular implementations of the required computational operations.

\begin{figure}
	\centering
	  \subfigure[]{\includegraphics[width=4cm]{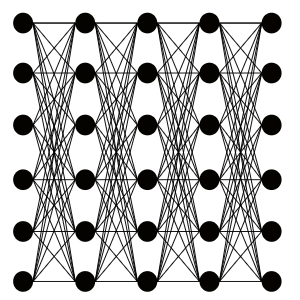}}
	  \hspace{1.5em}
	  \subfigure[]{\includegraphics[width=3.6cm]{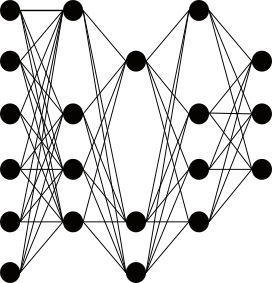}}
	  \hspace{1.5em}
	  \subfigure[]{\includegraphics[width=4cm]{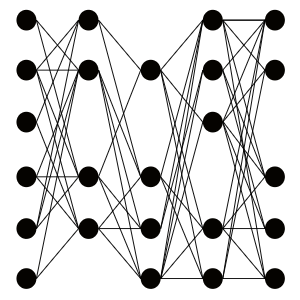}}
	\caption{A fully connected network (a), with sub-networks that are obtained by structured sparsity (node sparsity) (b), and unstructured sparsity (arbitrary node/connection sparsity) (c).}
	\label{fig:fig1}
\end{figure}

We distinguish between these two types of sparsity and refer to node sparsity as \textit{structured~sparsity} and the general case of sparsity by arbitrary node and connections as \textit{unstructured~sparsity}~(Figure~\ref{fig:fig1}). This definition is loosely aligned with the work of Lie et al. \cite{liu2018rethinking}. The results reported here involve sub-networks produced with structured~sparsity. In this sense, Weedout resulting in unstructured~sparsity would be similar to performing \textit{DropConnect} \cite{wan2013regularization} on a network, except that it is not applied during the training phase.

\begin{figure}
	\centering
	  \subfigure[]{\includegraphics[width=15cm]{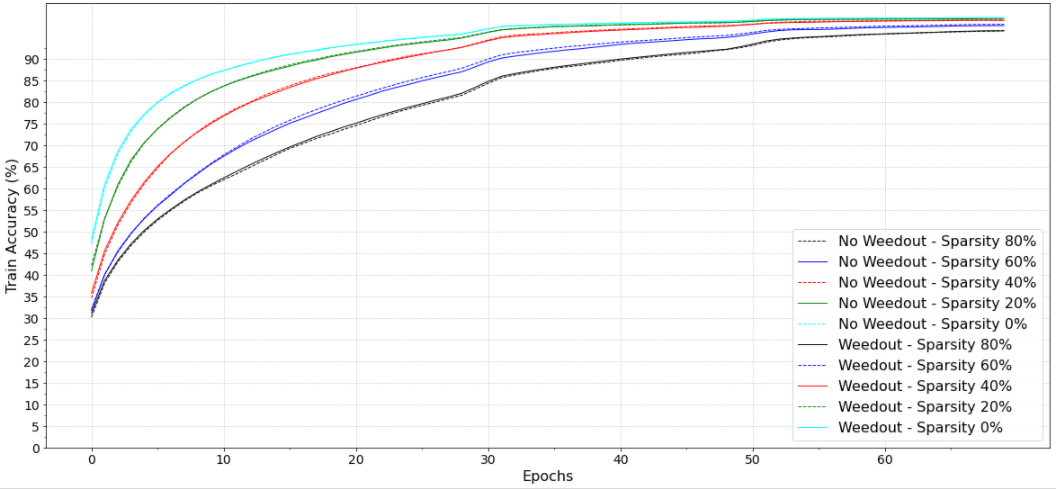}}
	  \hspace{1.5em}
	  \subfigure[]{\includegraphics[width=15cm]{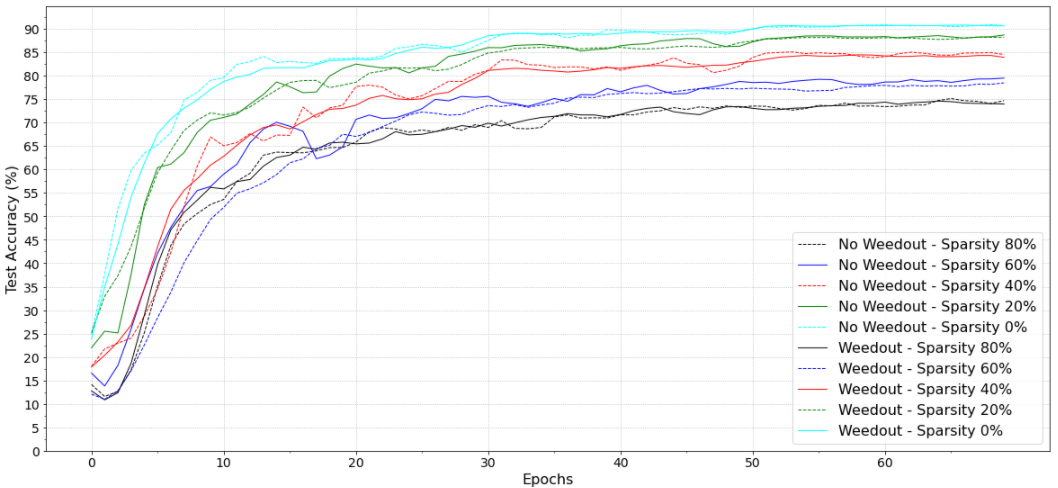}}
	\caption{Train accuracy (a) and test accuracy (b) progression.}
	\label{fig:fig2}
\end{figure}

We train a first set of Wide ResNet models at sparsity levels 0\%, 20\%, 40\%, 60\%, and 80\% with no Random Search Weedout. We then train a second set at the same sparsity levels but with the Weedout process applied with the mentioned hyperparameters above. This way, we can control for the effect of Random Search Weedout at multiple sparsity levels. Figures~\ref{fig:fig2} shows the training and test accuracy (mean of 5 runs) performances of the models at various sparsity levels with or without Weedout.

\section{Results and Discussion}
For these Wide ResNet architectures, we observed a progressive deterioration of prediction accuracy on the test set as sparsity increases as shown in Figure~\ref{fig:fig2}. Interestingly, there is only approximately 15\% drop in test accuracy between a fully dense network and one with 80\% sparsity. This can be a reasonable trade-off for applications where parameter efficiency is of critical importance. 

We also observed from the experiment that Random Search was capable of finding sparse configurations among the random population which performed better than others as measured by validation set accuracy. However, once gradient-based learning started and the training dataset was visited over multiple epochs, the best sparse configuration found by Random Search did not perform better than any other randomly initialized network at the same level of sparsity on the test set.

At multiple levels of sparsity, the random sparse networks reached the same level of performance as the Random Search found ones. Performance deterioration for a random configuration versus a Random Search found one is at the same pace and at the same level for a given degree of sparsity.

We conclude that while Random Search is able to find networks with better initialization, this advantage is quickly overrun by gradient-based learning over multiple passes on the training dataset. Hence any method that can successfully find an initial sparse configuration that performs better than a randomly initialized configuration post training is at least better than Random Search.

Finally, it should be noted that we examined Random Search as a baseline sparsity search method under very specific conditions. Our choice of task, architecture, and experiment parameters may be further studied for a more generalized and comprehensive view of its performance as a baseline. We briefly mention some points for future considerations below.

\section{Future Work}

\paragraph{Global sparsity}
in the current experiments, we controlled the sparsity level in the networks by removing a fixed portion of nodes from the network, per each layer. One can achieve the same level of sparsity, but from a global perspective, i.e. an arbitrary number of nodes from each layer would be removed to result in a specific level of global sparsity.

\paragraph{Unstructured sparsity:}
in general, the performance of structured versus unstructured sub-networks in overparameterized networks is not well understood in the context of pruning. While Liu et al. \cite{liu2018rethinking} have made some informative explorations in this area, there remains much opportunity to understand the task and computational performance differences between the two. This is especially important for choosing the right computational platform for deploying a sparse network.

\paragraph{Non-uniform sparsity:}
in our experiments we have applied Weedout to produce sparsity uniformly and layer-wise in all layers of the Wide ResNet convolutional neural network. Further experimentation may yield informative results on how non-uniform sparsity affects performance, e.g. a higher level of sparsity on lower-level feature extraction layers, versus less sparsity in higher representation space layers. Particularly for convolutional neural networks, sparsity can also be extended to channels, e.g. as done in some surveyed works \cite{liu2018rethinking, liu2017learning}.

\paragraph{Computational operations measurements:}
a more comprehensive picture of compute performance improvement due to sparsity search methods can be obtained by profiling and measuring total (e.g. including the Weedout phase) and per-phase computational operations and the training wall-time.

\paragraph{Degrees of overparameterization:}
we apply the Weedout technique to a Wide ResNet network, which is generally considered overparameterized on the CIFAR-10 task. However, further experiments on the effect of sparse network search at various degrees of overparameterization would result in a better empirical understanding of particular situations where search may be more effective. Possible next steps could involve the application of sparse network search to simpler tasks such as the MNIST digit classification and to harder tasks such as the CIFAR-100 image classification task for comparison.

\paragraph{RSH strategies and parameters:}
in the current experiments, we have only applied Random Search as Weedout's RSH strategy. Given the substantial evidence on the performance variation of different strategies and on different tasks \cite{DeJong_2016}, more sophisticated RSH algorithms may be applied in order to surface those that may be more effective for sub-network search in overparameterized networks. However, this will be out of the current scope of Random Search as a baseline method. 

We conducted experiments given a population of 100 randomly initialized networks, and ran Random Search for 5 generations. A more comprehensive assessment on the effect of population size and generational count may yield different results, i.e. larger populations or more generations may result in Random Search finding configurations that would continue to show better performance even post gradient-based training.

\paragraph{Weedout after limited training:}
so far, the fitness value for each sparse sub-network configuration has been the inference performance measurement of the network on a batch of validation data prior to any weight updates. Given that a substantial amount of learning in a network is skewed towards the early stages of training, one can experiment with fitness values obtained at varying stages of training steps and attempt to balance using a stronger search signal in the face of additional compute cost.

\paragraph{Weedout throughout the training process:}
the sparsity search process can be continued through the entire training process, i.e. similar to Dropout \cite{Srivastava_Hinton_Krizhevsky_Sutskever_Salakhutdinov_2014}, but where finding the Dropout binary mask is under search. We envision the opportunity of finding effective masks while continuing to update the network weights, and arriving at a mask which can be immediately used to prune the overparameterized network. Of course in this case, we lose the opportunity of finding an efficient sparse sub-network prior to training and gaining training compute performance efficiency.

\bibliographystyle{ieeetr}
\bibliography{references}

\newpage

\section*{Appendix}
The sets of figures below show the train and test accuracy performance of all the models at various sparsity levels with the 95\% confidence interval from 5 runs.

\begin{figure}[!htbp]
	\centering
	  \subfigure[]{\includegraphics[width=8cm]{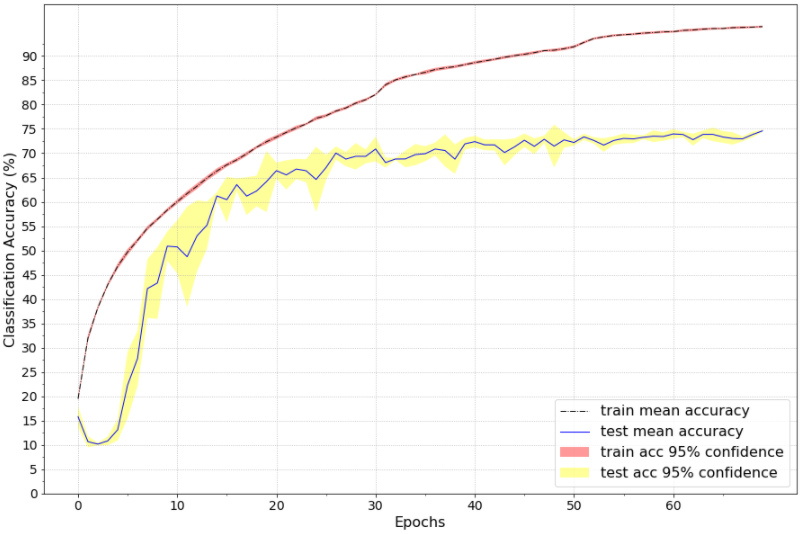}}
	  \hspace{0em}
	  \subfigure[]{\includegraphics[width=8cm]{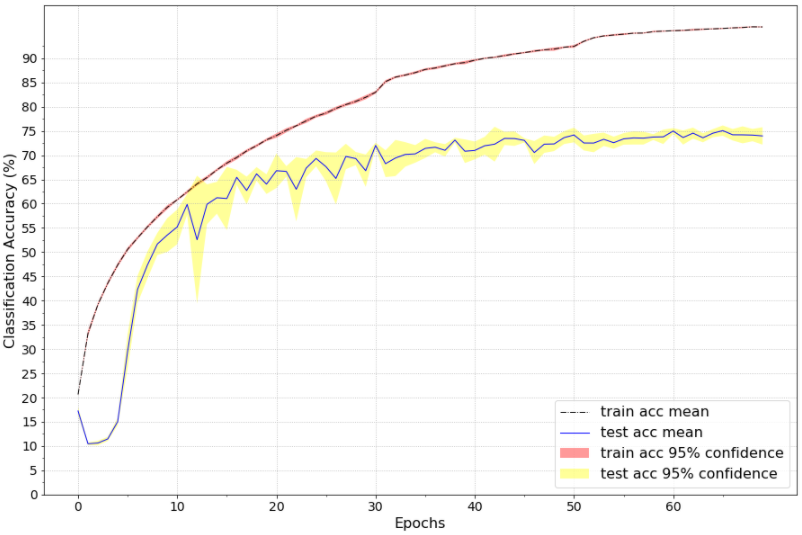}}
	  \hspace{0em}
	  \subfigure[]{\includegraphics[width=8cm]{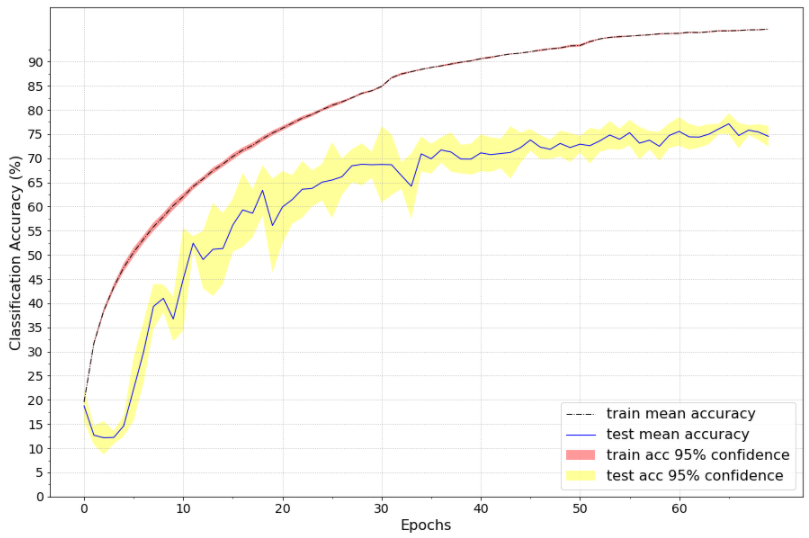}}
	  \hspace{0em}
	  \subfigure[]{\includegraphics[width=8cm]{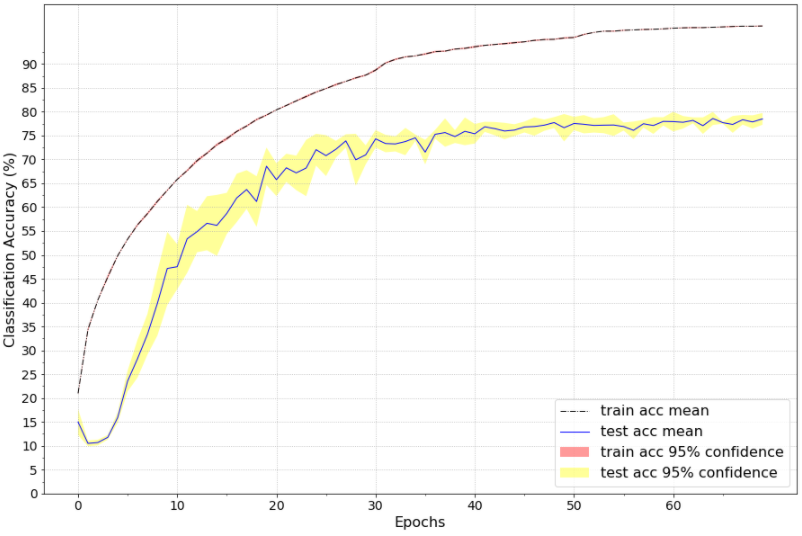}}
	  \hspace{0em}
	  \subfigure[]{\includegraphics[width=8cm]{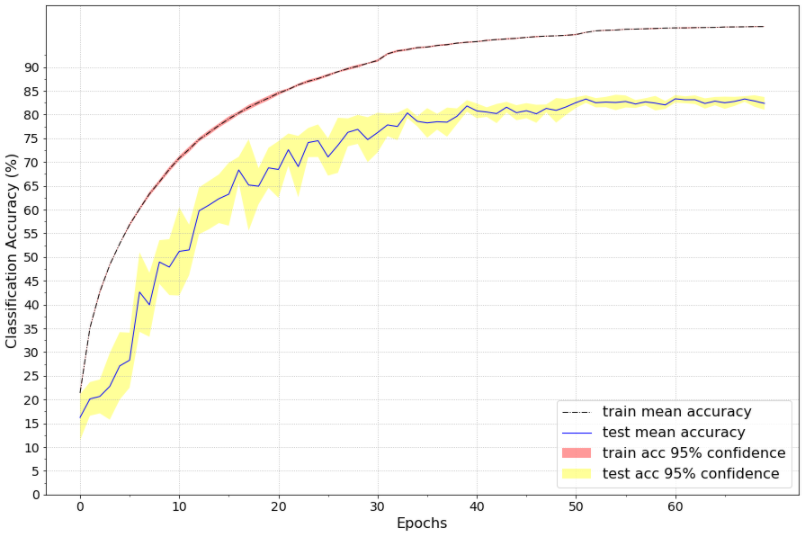}}
	  \hspace{0em}
	  \subfigure[]{\includegraphics[width=8cm]{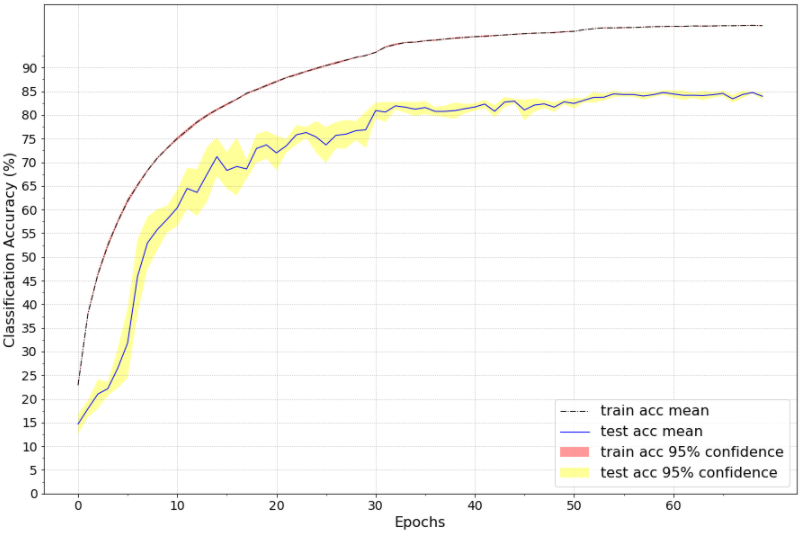}}
	  \hspace{0em}
\end{figure}

\FloatBarrier

\begin{figure}[!htbp]
    \ContinuedFloat
	\centering
	  \subfigure[]{\includegraphics[width=8cm]{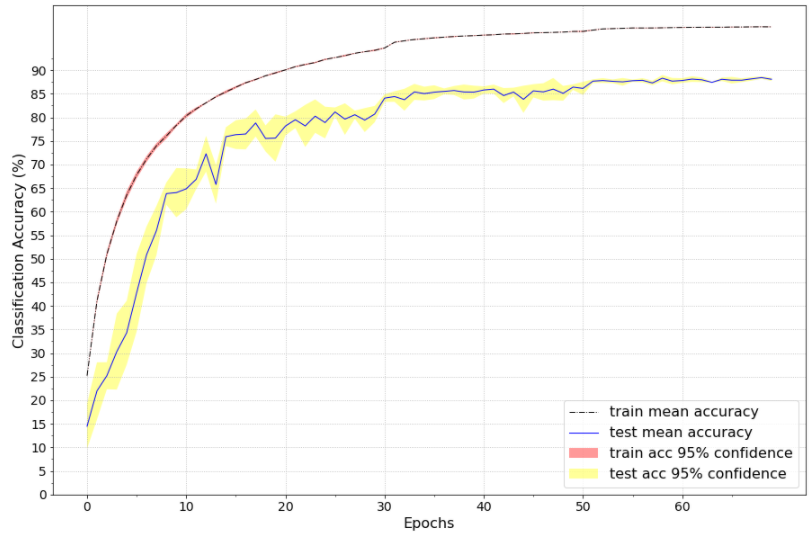}}
	  \hspace{0em}
	  \subfigure[]{\includegraphics[width=8cm]{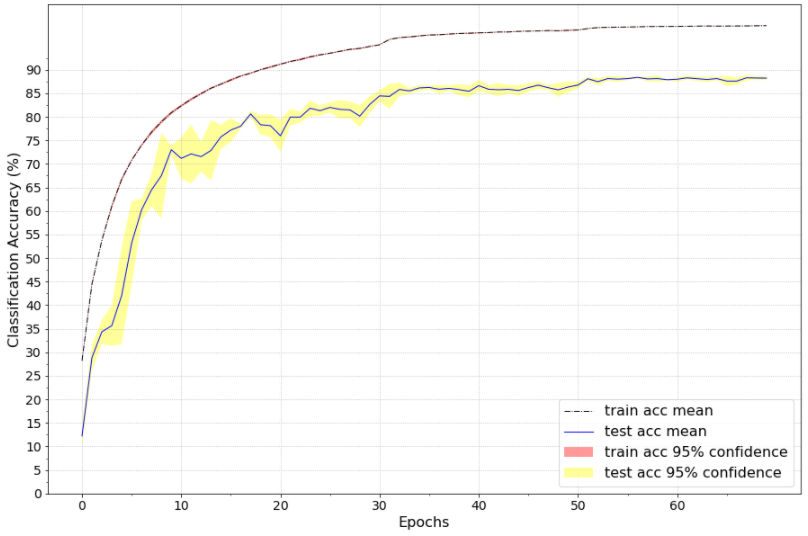}}
	  \hspace{0em}
	  \subfigure[]{\includegraphics[width=8cm]{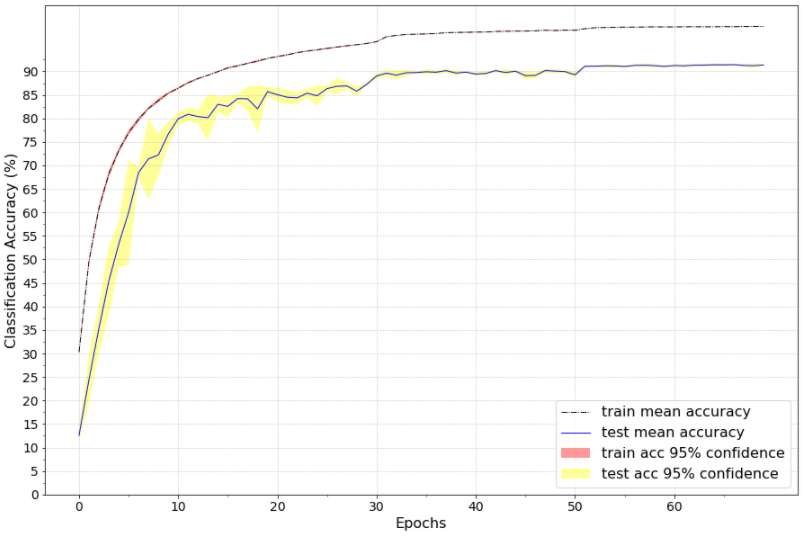}}
	  \hspace{0em}
	  \subfigure[]{\includegraphics[width=8cm]{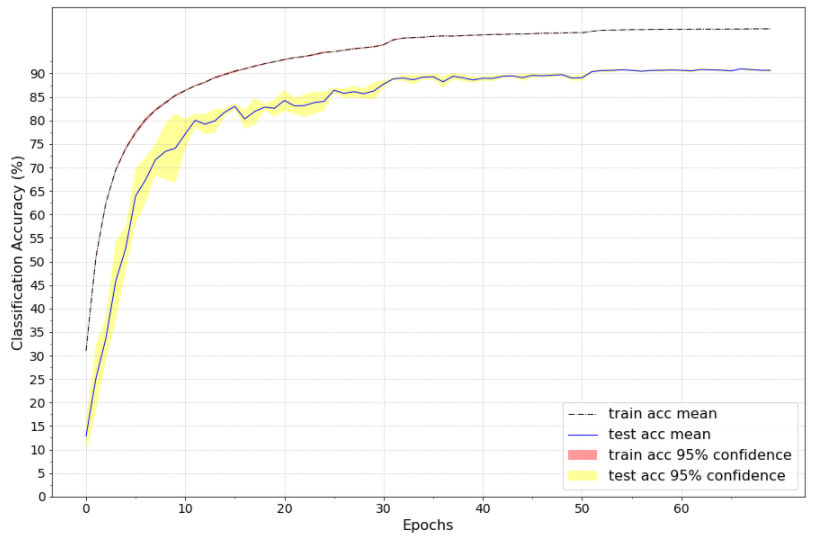}}
	\caption*{(a) No Weedout and sparsity at 80\%, (b) Weedout and sparsity at 80\%), (c) No Weedout and sparsity at 60\%, (d) Weedout and sparsity at 60\%, (e) No Weedout and sparsity at 40\%, (f) Weedout and sparsity at 40\%, (g) No Weedout and sparsity at 20\%, (h) Weedout and sparsity at 20\%, (i) No Weedout and sparsity at 0\%, (j) Weedout and sparsity at 0\%.}
\end{figure}

\end{document}